\documentclass{article}
\usepackage{spconf,amsmath,graphicx,booktabs}
\usepackage{amssymb}
\usepackage{adjustbox}
\usepackage{xcolor}
\usepackage{float}

\usepackage[pagebackref=true,breaklinks=true,letterpaper=true,colorlinks,bookmarks=false]{hyperref}

\usepackage[linesnumbered,ruled,boxed,norelsize,vlined,commentsnumbered]{algorithm2e}

\newcommand{\tg}[1]{\textcolor{green}{#1}}
\usepackage{array} 

\title{Image Augmentation with Controlled Diffusion for WEAKLY-SUPERVISED SEMANTIC SEGMENTATION}
\name{Wangyu Wu$^{1,2}$, Tianhong Dai$^{3}$, Xiaowei Huang$^{2}$, Fei Ma$^{1^{*}}$, Jimin Xiao$^{1^{*}}$\thanks{$^{*}$Corresponding authors}\thanks{The work was supported by National Key R\&D Program of China (No.2022Y
FE0200300), National Natural Science Foundation of China under 61972323, and XJTLU AI University Research Centre, Jiangsu Province Engineering Research Centre of Data Science and Cognitive Computation at XJTLU and SIP AI innovation platform (YZCXPT2022103).}}

\address{$^{1}$Xi’an Jiaotong-Liverpool University\quad
$^{2}$The University of Liverpool \quad $^{3}$University of Aberdeen
}
\begin{document}
\maketitle
\begin{abstract}
Weakly-supervised semantic segmentation (WSSS), which aims to train segmentation models solely using image-level labels, has achieved significant attention. Existing methods primarily focus on generating high-quality pseudo labels using available images and their image-level labels. However, the quality of pseudo labels degrades significantly when the size of available dataset is limited.  
Thus, in this paper, we tackle this problem from a different view by introducing a novel approach called \textit{Image Augmentation with Controlled Diffusion} (IACD). This framework effectively augments existing labeled datasets by generating diverse images through controlled diffusion, where the available images and image-level labels are served as the controlling information.  
Moreover, we also propose a high-quality image selection strategy to mitigate the potential noise introduced by the randomness of diffusion models. In the experiments, our proposed IACD approach clearly surpasses existing state-of-the-art methods. This effect is more obvious when the amount of available data is small, demonstrating the effectiveness of our method.



\end{abstract}
\begin{keywords}
weakly-supervised semantic segmentation, diffusion model, high-quality image selection
\end{keywords}
\section{Introduction}
\label{sec:intro}

Weakly-supervised semantic segmentation (WSSS) leverages image-level labels to generate pixel-level pseudo masks for training the the segmentation models. The primary challenge lies in enhancing the quality of generated pseudo-labels. Currently, most methods involve injecting more category information into the network or performing additional information learning on existing training data ~\cite{xie2022exploring, chang2020weakly}, such as sub-class distinctions~\cite{chang2020weakly} and adding category information to network~\cite{xie2022exploring}. Alternatively, efforts are directed towards optimizing network structures~\cite{rossetti2022max,xu2022multi,ru2022learning,wu2023topk} to better suit learning in weakly-supervised scenarios. However, the aforementioned methods are all constrained by the scale of the available training data.
\begin{figure}[t]
\centering
\includegraphics[width=0.7\linewidth]{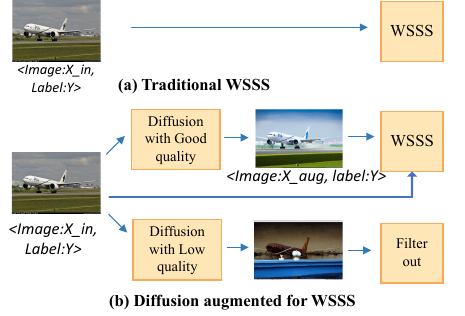}
\vspace{-0.3cm} 
\caption{(a) In the previous method, only images from the original dataset are used for training. (b) Our proposed IACD utilizes an diffusion model to generate synthetic images. Then, an image selection module is used to annotate and select the high-quality synthetic images to augment the original dataset for training.}
\label{fig:idea}
\vspace{-0.3cm} 
\end{figure}

\begin{figure*}[t] 
\vspace{-0.5cm}
\begin{center}
\includegraphics[width=0.8\linewidth]{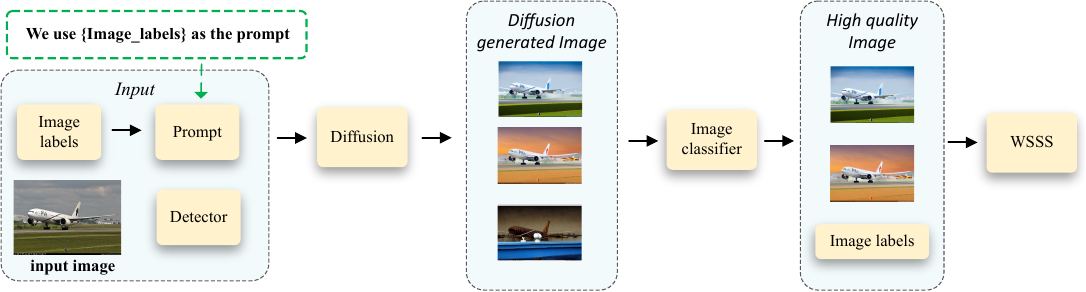}
\vskip -0.3cm
\caption{The pipeline of our IACD. The airplane in the showcase is using the diffusion model with prompts to generate candidate images. Subsequently, the candidate images are filtered through an image selection process to ensure that only high-quality images are used as training data for the downstream WSSS.}
\label{fig:framework}
\end{center}
\vspace{-0.3cm}
\end{figure*}
The Diffusion Probabilistic Model (DPM)~\cite{sohl2015deep} is an appealing choice for the aforementioned problem because it belongs to a class of deep generative models that have recently gained prominence in the field of computer vision~\cite{ho2020denoising,song2020denoising,song2020score}. The generated images exhibit high-quality with few artifacts and effectively align with the given text prompts, even when these prompts depict unrealistic scenarios that were not encountered during training. This highlights the robust generalization capabilities of diffusion models. Notably, recent works such as Stable Diffusion~\cite{rombach2022high} with ControlNet~\cite{zhang2023adding} is able to generate high-quality synthetic images. 

In this paper, we propose \textit{Image Augmentation with Controlled
Diffusion} (IACD) to generate high-quality synthetic training data for WSSS (see Fig.~\ref{fig:idea}). Our contributions are: 1). Our approach aims to enhance WSSS performance, which is the first proposal to utilize conditional diffusion for augmenting the original dataset with image-level labels. 2). An image selection approach is introduced, aiming to keep high-quality training data while effectively filtering out low-quality generated images. This strategy helps prevent any adverse impact on model training. 3). Our proposed framework outperforms all current state-of-the-art methods, and it shows varying performance improvements across different training data sizes, particularly at 5\% of the training data, where there is 4.9\% increase for the segmentation task on the validation set on the PASCAL VOC 2012 dataset.

\section{Methodology}
\label{sec:method}
In this section, we present the general framework and key
components of our proposed method. Initially, We introduce the overall architecture and pipeline of our IACD method (Sec.~\ref{sec:overall}). Then, a diffusion model based approach is proposed for data augmentation in WSSS tasks (Sec.~\ref{sec:DA}). Furthermore, we develop a high-quality image selection strategy that aims to ensure the quality of data generated by diffusion model, thereby reducing the model noise (Sec.~\ref{sec:im}). Finally, the components of the final dataset used for training are also discussed (Sec.~\ref{sec:finalinput}).


\subsection{Overall Structure} \label{sec:overall}
As illustrated in Fig.~\ref{fig:framework}, we utilize the diffusion model~\cite{rombach2022high} along with ControlNet~\cite{zhang2023adding} to generate new training samples under the guidance of conditioning inputs: original images and label prompts. In addition, we train a Vision Transformer (ViT) based image classifier~\cite{dosovitskiy2020image} by using existing dataset with image-level labels to select high-quality generated training samples. During the selection, we select generated images with high prediction scores as high-quality samples and filter out low-quality generated images with noise. Finally, we extend the original dataset with the generated samples for the training of WSSS.


\subsection{Controlled Diffusion for Data Augmentation} \label{sec:DA}
\begin{figure*}[t] 
\vspace{-0.5cm}
\begin{center}
\includegraphics[width=0.9\linewidth]{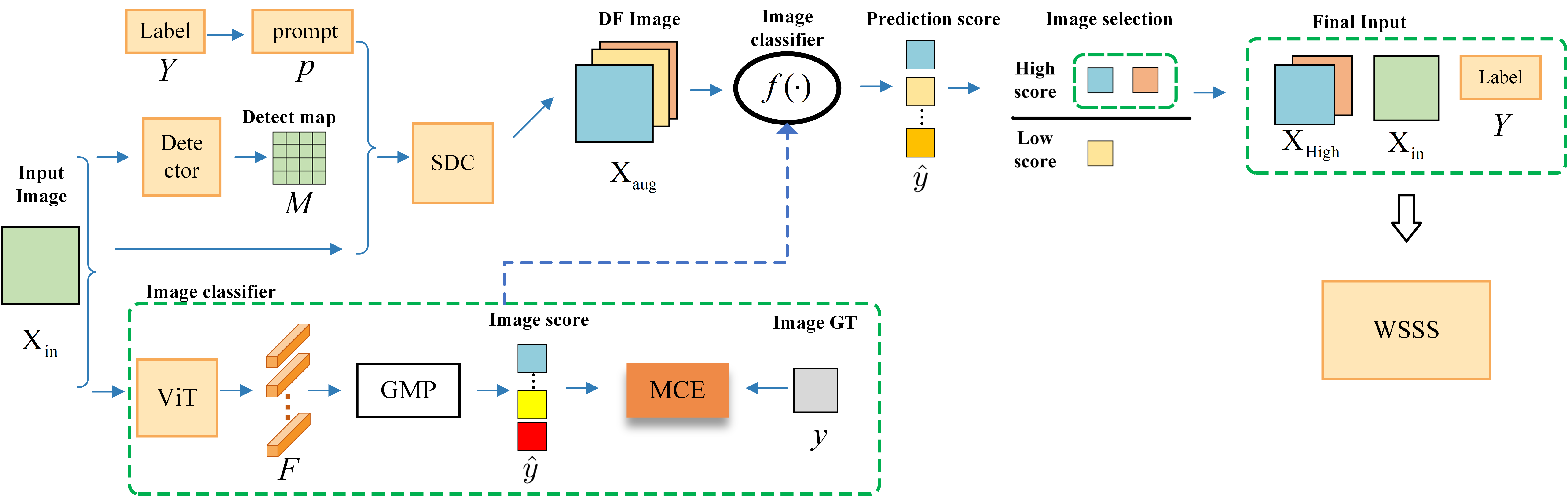}
\vskip -0.3cm
\caption{The overall framework of IACD consists of several steps. Firstly, IACD utilizes controlled diffusion to generate entirely different images. Subsequently, the original image is processed using the Vision Transformer (ViT) as an encoder to generate patch embeddings, and a patch-driven classifier is trained for image categorization. Then, the generated diffusion images are passed through the same trained image classifier to select a high-quality image set. Moreover, the selected image set, along with the original images and their corresponding labels, is passed to the downstream WSSS task.}
\label{fig:IACD}
\end{center}
\vspace{-0.4cm}
\end{figure*}

The motivation for using controlled image diffusion models to augment data is that these models can generate infinite and diverse task-specific synthetic images based on a given image and a text prompt. In this work, we utilize Stable Diffusion with ControlNet (SDC)~\cite{zhang2023adding} as our generative model (see Fig.~\ref{fig:IACD}). In the data augmentation stage, an input image $X_{in}\in\mathbb{R}^{h\times w\times 3}$, a text prompt $P$, and a detection map $M$ are feed into SDC $\delta(\cdot)$ to generate a new training data $X_{aug}$. More specifically, the text prompt is formulated by the corresponding image-level label $Y$. The detection map is an extra condition (\textit{e.g}., Canny Edge~\cite{4767851} and Openpose~\cite{8765346}) to control the generation results.

\begin{equation}
    X_{aug} = \delta(X_{in}, M, P).
\end{equation}
More details about the data augmentation process are described in Algo.~\ref{alg:CD}.

\begin{algorithm}
\caption{{\small{Diffusion Model for Data Augmentation}}}\label{alg:CD}
\KwIn{\parbox[t]{8cm}{
    an input image $X_{in}$, an image-level label $Y$}}
\KwOut{a generated image $X_{aug}$}
$P \gets \text{generate\_prompt}(Y)$\\
\eIf{$``person" \in Y$}{
$M \gets \text{detect\_map}(X_{in}, \text{human\_pose})$}
{$M \gets \text{detect\_map}(X_{in}, \text{canny\_edge})$}
$X_{aug} \gets \delta(X_{in}, M, P)$
%
\end{algorithm}



\subsection{High-quality Synthetic Image Selection} \label{sec:im}
In order to guarantee the quality of synthetic data that will be used for training, a selection strategy is introduced to select the high-quality generated samples. As shown in Fig.~\ref{fig:IACD}, a ViT-based patch-driven classifier is first trained by using the original dataset with the image-level labels. To train the classifier, the input image $X_{in}$ is divided into $s$ input patches  $X_{patch}\in \mathbb{R}^{d\times d \times 3}$ with the fixed size, where $s = \frac{hw}{d^{2}}$. Then, the patch embedding $F\in\mathbb{R}^{s\times e}$ is achieved by using ViT encoder. Next, a weight $W\in\mathbb{R}^{e\times|\mathcal{C}|}$ and a softmax function is applied to output the prediction scores $Z\in\mathbb{R}^{s\times|\mathcal{C}|}$of each patch:
\begin{equation}
    Z = \text{softmax}(FW),
\end{equation}
where $\mathcal{C}$ is the set of categories in the dataset. Global maximum pooling (GMP) is then used to select the highest prediction scores $\hat{y}\in\mathbb{R}^{1\times|\mathcal{C}|}$ in each class among all patches. Finally, $\hat{y}$ is utilized as the prediction scores for the image-level classification and the classifier is trained via using the multi-label classification prediction error (MCE):
\begin{equation} 
\begin{aligned}\label{eq:MCE}
\mathcal{L}_{MCE}&=\frac{1}{|\mathcal{C}|}\sum_{c\in\mathcal{C}}{BCE(y_c,\hat{y}_c)}\\
&=-\frac{1}{\mathcal{|C|}}\sum_{c\in\mathcal{C}}{y_c\log(\hat{y}_c)+(1-y_c)\log(1-\hat{y}_c)},
\end{aligned}
\end{equation}
where, $\hat{y}_{c}$ is the prediction score of class $c$ and $y_{c}$ is the ground-truth label. Once the classifier finishes the training, we can use it to select the high-quality generated training data.

In the selection stage, the synthetic image $X_{aug}$ generated by $\langle X_{in}, Y\rangle$, is passed into the classifier, followed by the GMP to output the image-level prediction score $\hat{y}$. Then, the classes with the scores above a certain threshold $\epsilon$ are used as the ground-truth label for the generated image: $Y_{aug}=\{c | \hat{y}_{c} > \epsilon\}$. If $Y_{aug}$ is a subset of the label of the input image $Y$, the generated sample $\langle X_{aug}, Y_{aug}\rangle$ will be added into the synthetic dataset $\mathcal{D}_{aug}$. More details about the image selection are described in Algo.~\ref{alg:HQIS}.

\begin{algorithm}
\caption{{\small{High-Quality Image Selection}}}\label{alg:HQIS}
\KwIn{\parbox[t]{7cm}{
        a ground-truth label of the input image $Y$, a generated image $X_{aug}$, a prediction score $\hat{y}$, a label of generated image $Y_{aug}$, a set of classes $\mathcal{C}$, a threshold $\epsilon$ and a synthetic dataset $\mathcal{D}_{aug}$
    }}
\KwOut{a synthetic dataset $\mathcal{D}_{aug}$}
 $Y_{aug} \gets \varnothing$\\
 \ForEach{$c\in \mathcal{C}$}{ 
\If{$\hat{y}_{c} > \epsilon$}{
    $Y_{aug} \gets Y_{aug} \cup \{c\}$
}
}
\If{$Y_{aug} \subseteq Y$} {
    $\mathcal{D}_{aug} \gets \mathcal{D}_{aug} \cup \{\langle X_{aug}, Y_{aug}\rangle\}$
}
\end{algorithm}


This selection strategy serves two purposes. First, it only keeps the synthetic samples with high prediction scores in specific categories, which guarantees there is a high probability that objects of these classes will appear in the synthetic image. Second, the synthetic image will not contain objects that do not belong to the image-level label of the input image. In this way, the quality of synthetic dataset $\mathcal{D}_{aug}$ can be improved.
\subsection{Final Training Dataset of WSSS} \label{sec:finalinput}
After selecting the high-quality generated training samples, the synthetic dataset $\mathcal{D}_{aug}$ and the original dataset $\mathcal{D}_{origin}$ are combined as an extended dataset $\mathcal{D}_{final}$ for the training of WSSS: $\mathcal{D}_{final} = \mathcal{D}_{origin} \cup \mathcal{D}_{aug}$.  

\section{Experiments}
\label{sec:Experiments}

In this section, we describe the experimental setup, including dataset, evaluation metrics, and implementation details. We then compare our method with state-of-the-art approaches on PASCAL VOC 2012~\cite{everingham2010pascal}. Finally, ablation studies are performed to validate the effectiveness of the proposed method.

\subsection{Experimental Settings}

\textbf{Dataset and Evaluated Metric.} We conduct our experiments on PASCAL VOC 2012~\cite{everingham2010pascal}, which comprises $21$ categories, including the additional background class. The PASCAL VOC 2012 Dataset is typically augmented with the SBD dataset~\cite{hariharan2011semantic}. In total, we utilize 10,582 images with image-level annotations for training, and 1,449 images for validation. The training sets of Pascal VOC contain images with only image-level labels. We report the mean Intersection-Over-Union (mIoU) as the evaluation criterion. Additionally, we also evaluate the performance of our IACD method when the amount of original training data is gradually reduced from 10,582 (100\%) to 529 (5\%).

\noindent\textbf{Implementation Details.} In our experiments, we employe the ViT-B/16 as the ViT model, and we use the stable diffusion model~\cite{rombach2022high} with ControlNet~\cite{zhang2023adding} as our diffusion model. Images are resized to 384×384 pixels~\cite{kolesnikov2016seed} during the training of the patch-driven image classifier. The 24×24 encoded patch features are retained as input. The model is trained with a batch size of 16 for a maximum of 80 epochs. The image selection threshold $\epsilon$ is 0.9. We use Canny Edge~\cite{4767851} and Openpose~\cite{8765346} as detectors for ControlNet~\cite{zhang2023adding}, with a total of 20 diffusion steps.. Due to limitations in computational resources, we generate additional 10,582 images using a diffusion model in the experiments. During the WSSS training stage, we combine our synthetic dataset with the original training dataset as our final training dataset. Subsequently, we selecte ViT-PCM ~\cite{rossetti2022max} as our WSSS framework without any modifications. Our final training dataset serve as input for the WSSS framework, while keeping all other settings consistent with ViT-PCM~\cite{rossetti2022max}. The experiments are conducted using two NVIDIA 4090 GPUs. Finally, we use the same verification task and settings as ViT-PCM~\cite{rossetti2022max}.


\begin{table}[h!]
\centering
\caption{The comparison of segmentation performance on different sizes of training data.}
\label{tab:small}
\begin{adjustbox}{width=\linewidth}
\begin{tabular}{ccccccc}
\toprule
Percentage of Train Data & Baseline on Val & IACD on Val \\
\midrule
5\% &62.6\%&\textbf{67.5\% \tg{+4.9\%}} \\
15\% &65.6\%& \textbf{68.5\% \tg{+2.9\%}} \\
50\% &68.2\%& \textbf{70.5\% \tg{+2.3\%}} \\
100\% & 69.3\% &  \textbf{71.4\% \tg{+2.1\%}} \\

\bottomrule
\end{tabular}
\end{adjustbox}
\vspace{-1.8em}
\end{table}

\begin{table}[h!]
\centering
\caption{The comparison of semantic segmentation performance by using only pseudo masks for training.}
\label{tab:vocseg}
\begin{adjustbox}{width=0.9\linewidth}
\begin{tabular}{@{}ccccc@{}}

\toprule
Percentage of Train Data & Model & Pub. & mIoU (\%)\\
\midrule
100\%&MCTformer~\cite{xu2022multi} & CVPR22 & 61.7 \\
100\%&PPC~\cite{du2022weakly} & CVPR22 & 61.5\\
100\%&SIPE~\cite{chen2022self} & CVPR22 & 58.6\\
100\%&AFA~\cite{ru2022learning} & CVPR22 & 63.8 \\
100\%&ViT-PCM~\cite{dosovitskiy2020image} & ECCV22 & 69.3 \\
\midrule
\textbf{50\% }& \textbf{IACD (Ours) + ViT-PCM} &  & \textbf{70.5 } \\
\textbf{100\% } &\textbf{IACD (Ours) + ViT-PCM} &  & \textbf{71.4 } \\

\bottomrule
\end{tabular}
\end{adjustbox}
\vspace{-3mm}
\end{table}

\begin{table}[h!]
\centering
\caption{Ablation study on the data augmentation module and the high-quality image selection module.}
\label{tab:ablation}
\begin{adjustbox}{width=\linewidth}
\begin{tabular}{ccccccc}
\toprule
Backbone & Original Train & Diffusion Augmentation & Image Selection & Result on Val \\
\midrule

ViT-B/16 &\checkmark &  &  & 69.3\% \\
ViT-B/16 &\checkmark &  \checkmark & & 69.1\% \\
ViT-B/16 &\checkmark &  \checkmark & \checkmark & 71.4\% \\

\bottomrule
\end{tabular}
\end{adjustbox}
\vspace{-1.8em}
\end{table}
\subsection{Comparison with State-of-the-arts}
\textbf{Comparison of Different Data Percentage.} Our proposed IACD method effectively enhances the original training dataset size in Tab.~\ref{tab:small}. As part of upstream data augmentation, it greatly aids the downstream WSSS framework in achieving higher segmentation performance. Furthermore, we observed that when the amount of training data is smaller, the effect is more obvious. This suggests that our approach is highly effective to augment the dataset and improve the performance when the amount of training data is insufficient.

\begin{figure}[t]
\centering
\includegraphics[width=\linewidth]{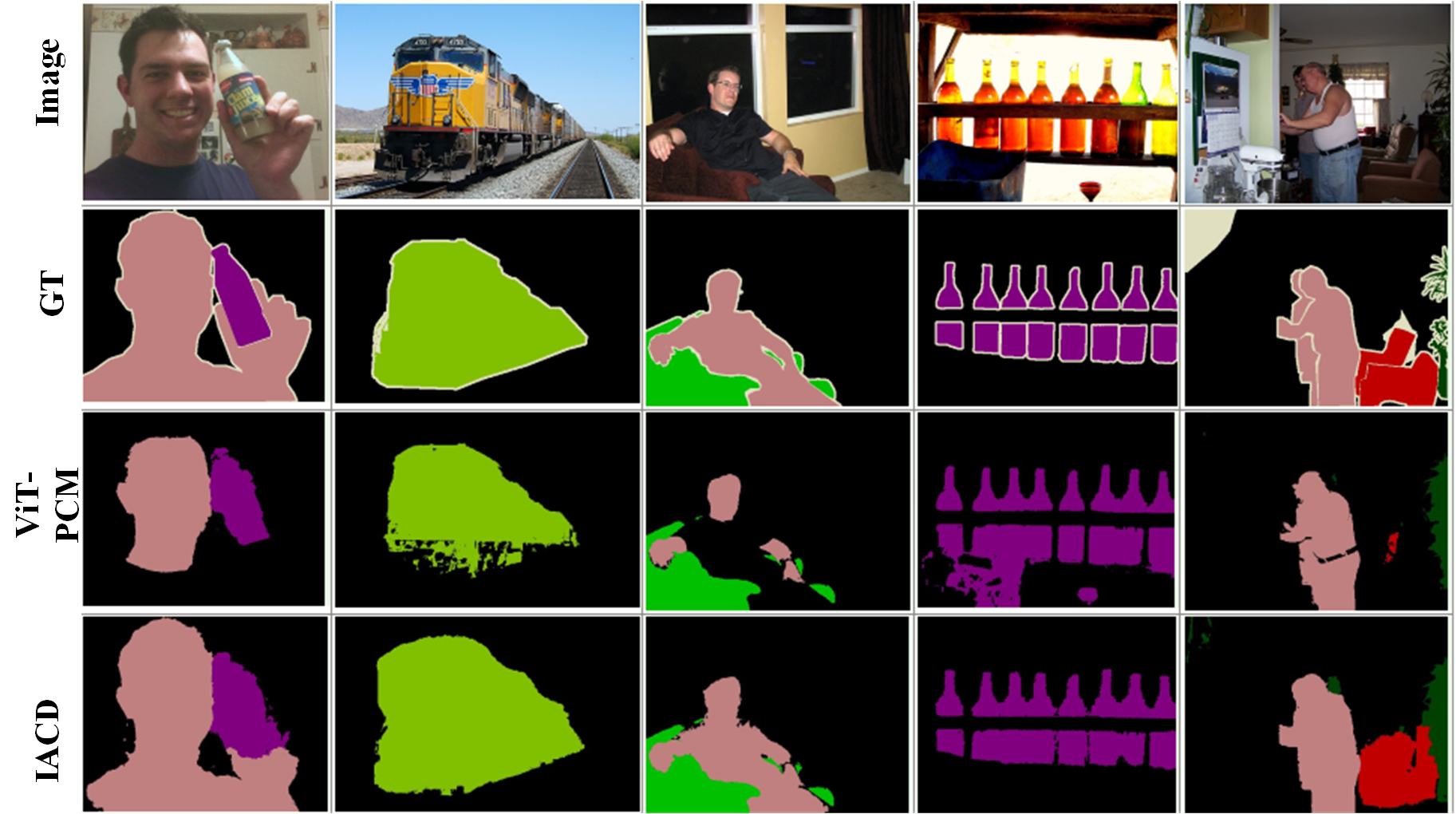}
\vspace{-2em} 
\caption{The comparison of qualitative segmentation results with ViT-PCM~\cite{dosovitskiy2020image}.}
\label{fig:result}
\vspace{-1em} 
\end{figure}

\noindent\textbf{Improvements in Segmentation Results.} To assess our methods, we apply our approach to the current state-of-the-art ViT-PCM as upstream data augmentation, while keeping the downstream WSSS consistent with the existing ViT-PCM. We then compare the segmentation results with the state-of-the-art techniques in Tab.~\ref{tab:vocseg}. Even with only 50\% of the train data, our method outperforms the baseline method ViT-PCM~\cite{dosovitskiy2020image}. 
The comparison of qualitative segmentation results are shown in Fig.~\ref{fig:result}.

\subsection{Ablation Studies}

We conducted an ablation study to assess the impact of our two key contributions: diffusion augmentation and high-quality image selection. As shown in Tab.~\ref{tab:ablation}, our diffusion augmentation introduces some random noisy images generated by the diffusion model, resulting in a 0.2\% decrease in mIoU on the validation set. Additionally, the proposed high-quality image selection effectively reduces noisy images by filtering out low-quality ones, leading to a 2.1\% improvement in mIoU for the baseline WSSS framework. When these two methods are combined, our comprehensive approach significantly outperforms the original framework.


\section{Conclusion}
\label{sec:conclusion}

In this work, we propose the IACD approach for data augmentation in weakly supervised semantic segmentation (WSSS). Unlike previous methods that focus on optimizing network structures or mining information from existing images, we introduce a diffusion model based module to augment additional data for training. To guarantee the quality of generate images, a high-quality image selection module is also proposed. By combining these two components, our approach has better performance than other state-of-the-art methods on PASCAL VOC 2012 dataset.

\bibliographystyle{IEEEbib}
\bibliography{refs}

\end{document}